\def\year{2021}\relax
\documentclass[letterpaper]{article} 
\usepackage{aaai21}  
\usepackage{times}  
\usepackage{helvet} 
\usepackage{courier}  
\usepackage[hyphens]{url}  
\usepackage{graphicx} 
\usepackage{natbib}  
\usepackage{caption} 
\usepackage{xcolor}

\title{Semantic Journeys:\\Quantifying Change in Emoji Meaning from 2012--2018}
\author {
    Alexander Robertson,\textsuperscript{\rm 1}
    Farhana Ferdousi Liza,\textsuperscript{\rm 2} \\
    Dong Nguyen,\textsuperscript{\rm 3}
    Barbara McGillivray,\textsuperscript{\rm 4,6}
    Scott A.\ Hale\textsuperscript{\rm 5,6}\\
}
\affiliations {
    \textsuperscript{\rm 1}University of Edinburgh 
    \textsuperscript{\rm 2}University of Essex 
    \textsuperscript{\rm 3}Utrecht University \\
    \textsuperscript{\rm 4}University of Cambridge 
    \textsuperscript{\rm 5}University of Oxford 
    \textsuperscript{\rm 6}Alan Turing Institute \\
    
    alexander.robertson@ed.ac.uk,
    farhana.ferdousi.liza@essex.ac.uk, \\
    d.p.nguyen@uu.nl, 
    bmcgillivray@turing.ac.uk,
    scott.hale@oii.ox.ac.uk
}

\newcommand{\emoji}[1]{\includegraphics[height=1.40\fontcharht\font`\A]{emoji_images/#1.png}}
\newcommand{\alex}[1]{\textcolor{black}{#1}}

\usepackage{booktabs}
\usepackage{varioref}
\usepackage{cleveref}

\begin{document}

\maketitle

\begin{abstract}

The semantics of emoji has, to date, been considered from a static perspective. We offer the first longitudinal study of how emoji semantics changes over time, applying techniques from computational linguistics to six years of Twitter data. We identify five patterns in emoji semantic development and find evidence that the less abstract an emoji is, the more likely it is to undergo semantic change. In addition, we analyse select emoji in more detail, examining the effect of seasonality and world events on emoji semantics. To aid future work on emoji and semantics, we make our data publicly available along with a web-based interface that anyone can use to explore semantic change in emoji.






\end{abstract}

\section{Introduction}\label{sec:intro}



Semantic change is the development of a word's meaning over time, e.g. \emph{cute} (\emph{shrewd} to \emph{pleasant}) and \emph{gay} (\emph{bright} to \emph{homosexual}). Such changes are only known to us because instances of both usage can be found in the historical record. It is entirely possible that there are other shifts and meanings of which we have no knowledge, purely because we have no direct evidence. This lack of evidence may be due to the loss of historical documents, or because the speech community it was found in did not record it before themselves dying out.

Emoji offer a unique perspective on this. Unlike words, we know exactly when each individual emoji came into existence and are pre-warned of new ones. The origin of words is far less clear, with new coinings and neologisms difficult to accurately detect. We also have unparalleled access to instances of emoji usage, through social media. We find ourselves in the hitherto unimaginable situation of being able to observe semantic change directly, from the birth of a term right up to the present moment.

Perhaps owing to their newness, prior work on emoji has taken a synchronic approach---studying emoji in general with no regard to time, implicitly assuming that emoji today are the same as ten days or years ago. We depart from this and present a diachronic view of emoji semantics, examining how the meaning of individual emoji evolves in six years of Twitter data. This approach is motivated by the development of techniques for semantic change detection, which so far have been applied only to words.

\alex{The main contribution of this work is the application of semantic change detection techniques to emoji and} we present the first longitudinal analysis of emoji semantics. We examine semantic development over time and search for patterns in this development. Although our analyses are mainly exploratory, we also analyse in more detail some of the patterns we observe. We conclude with case studies of a few select emoji, presenting their semantic journeys in more detail and account for their specific patterns in more detail by relating them to world events. 

\alex{Our finding that emoji can undergo semantic change over time will be of interest to linguists, and may motivate the application of linguistic theories of semantic change and development to emoji. They also highlight the need to consider semantic variation and change in emoji when designing NLP systems which include emoji---emoji semantics are not as static and simplistic as might be assumed.} To aid future work, we make our data publicly available and release a website\footnote{https://semantic-change.emoji-research.com - see Appendix A for more details.} where anyone can explore emoji semantic change. 


\section{Previous Work}

There is a growing literature on the semantics and pragmatics of emoji. Core themes are the sentiment associated with emoji \citep{novak2015sentiment}; how individual emoji are interpreted by readers, with context \citep{miller2016blissfully} or without context \citep{miller2017understanding}; how people use new emoji features like skin-tone modifiers \citep{robertson2018,robertson2020}; how particular emoji affect the interpretation of messages \citep{tigwell2016oh}; and how these findings compare to emoji-like constructs such as Animoji \citep{animoji}. There has been less work on the syntax of emoji and though there is evidence of an emergent syntax \citep{emojiwordorder}, it is emoji's semantic and pragmatic properties which are most suggestive of them being, if not a language, at least language-like.

Research into computational methods for the automatic detection of semantic change is currently an active area of research. Researchers have proposed a range of different approaches, from topic-based models \cite{cook2014novel,lau2014learning,frermann2016bayesian}, to graph-based models \cite{mitra2015automatic,tahmasebi2017finding}, and word embeddings \cite{Kim14,Basile2018,kulkarni2015statistically,hamilton2016diachronic,dubossarsky2017outta,tahmasebi2018study,Dubossarskyetal19}. \citet{ArXiV-Tahmasebi18} and \citet{kutuzov-etal-2018-diachronic} provide good overviews of this field and \citet{schlechtweg-etal-2020-semeval} report on the results of the first SemEval shared task on semantic change detection. The most successful methods are based on word embeddings, and most work has focused on semantic change over a relatively long time span, typically a few centuries \cite{hamilton-etal-2016-cultural, perrone_2019, schlechtweg-etal-2020-semeval}, with some notable exceptions, such as  \citet{shoemarketal2019}, who perform a systematic evaluation of embedding-based methods for short-term semantic change detection using Twitter data from 2011 to 2017. All methods proposed so far have dealt with semantic change of words, and to our knowledge no work has yet been done on applying this research to emoji specifically.

Emoji are generally considered as time-invariant artefacts, exhibiting variation only between each other or between populations of emoji users: i.e. iOS versus Android \citep{millercrossplatform}, young versus old \citep{herring2020gender}, male versus female \citep{chen2017through}. However, there is some work on how small close-knit groups apply their own special meaning to specific emoji \citep{Wisemanemojiaffordance}. 

\section{Data}

We start with the publicly available dataset of monthly Twitter embeddings created by \citet{shoemarketal2019}, who used them to evaluate different embedding and semantic change detection methods. Summary statistics are shown in Table~\ref{table:summarystats}. 

\begin{table}[]
\centering
\resizebox{\linewidth}{!}{%
\begin{tabular}{rlcccc}
\toprule
\multicolumn{1}{l}{} &  & \multicolumn{1}{l}{} & \multicolumn{3}{l}{Monthly embeddings} \\ \cline{4-6}
\multicolumn{1}{c}{Version} & \multicolumn{1}{c}{Date} & Emoji & median & min & max \\ \midrule
unicode-6.0 & Oct 2011 & 494 & 57 & 1 & 66 \\
unicode-6.1 & Jan 2012 & 13 & 58 & 58 & 58 \\
unicode-7.0 & Jun 2014 & 40 & 16 & 1 & 26 \\
unicode-8.0 & Jun 2015 & 11 & 22 & 4 & 25 \\
emoji-3.0 & Jun 2016 & 3 & 10 & 6 & 10 \\ \bottomrule
\end{tabular}%
}
\caption{Emoji per release, with summary of number of monthly embeddings per emoji.}
\label{table:summarystats}
\end{table}

Due to sporadic data collection issues, we do not have complete data for some months. These are therefore excluded. Due to tokenisation issues\footnote{Which are understandable, given \citet{shoemarketal2019} focused on words, not emoji.}, we can retain only those emoji which are only found as a single Unicode sequence. For example, \emoji{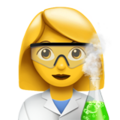} is actually composed of \emoji{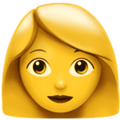} and \emoji{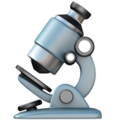}. However, \emoji{microscope} can appear on its own, so its embedding will have been trained on \emoji{microscope} and \emoji{woman-scientist}. This risks clouding the semantics encoded by the embedding. This issue affected 45 single codepoint emoji in total, plus all sequences they appear in. We have no data for Emoji 1.0 or 2.0 releases and very little for 3.0---these contain mainly multi-codepoint emoji sequences.
 
The first month for which we have data is January 2012. As emoji were only widely available outside Japan from October 2011, when Apple released iOS 5, the trained embeddings very nearly capture the ``birth'' of many Unicode 6.0 and 6.1 emoji and we have an almost complete timeline of embeddings for most of these. In our analyses, we therefore focus on Unicode 6.x emoji but do include all emoji in our first explorations of the data.

\section{Methods}

The aim of this paper is to characterise the development of emoji semantics from 2012 to 2018. We apply methods from prior work on words' semantic change. Due to the large number of emoji we examine, we use unsupervised clustering techniques to automatically identify patterns within the developmental trajectories we extract. In this section, we provide the technical details of our techniques.

\subsection{Measuring semantic change}\label{methods:scc}

We measure semantic change at each month of an emoji's existence, following the local neighbourhood measure of semantic change described by \citet{hamilton-etal-2016-cultural}. This technique has been widely used, for example to track semantic change over time in gender/ethnic stereotypes \cite{garg2018word} or trace the dynamics of global armed conflicts \cite{kutuzov2017tracing}. Crucially, it has also been used to track semantic change on social media and found to perform well in that context \cite{shoemarketal2019}.

The local neighbourhood measure of semantic change works as follows. For a target word (or emoji, in our case), the first available monthly embedding is set as an ``anchor'' as per \citet{shoemarketal2019}. All subsequent monthly embeddings are compared against this. The comparison involves constructing two ``second-order'' vectors for each embedding, whose components are the cosine similarities between the target word and its $k$ nearest neighbours. The cosine similarity between these two vectors represents the local neighbourhood change for that timestep. By always calculating semantic change scores relative to the second-order vector generated from a word's first appearance (its anchor point), we can measure that item's semantic change score over time. From here on, we refer to this as the semantic change (SC) score.

We use 25 neighbours per second-order vector (as per \citet{hamilton-etal-2016-cultural}). We exclude hashtags from the list of nearest neighbours as they are generally ephemeral, linked to specific events or trends and serve a variety of platform-specific purposes \citep{hashtagdiffusion, whatistwitter, hashtagtemporal}. We also exclude emoji, to better understand emoji relative to words. The 25 nearest neighbours are taken from a filtered selection of the 500 nearest neighbours\footnote{Because we require each neighbour to appear in both time steps, a second-order vector may have fewer than 25 components because vocabularies change at each time step. In practice this affects $\sim$2,000 of the $\sim$25,000 vector pairs we construct and 90\% of these have at least 15 components.}.

\begin{figure*} 
    \centering
    \includegraphics[width=\textwidth]{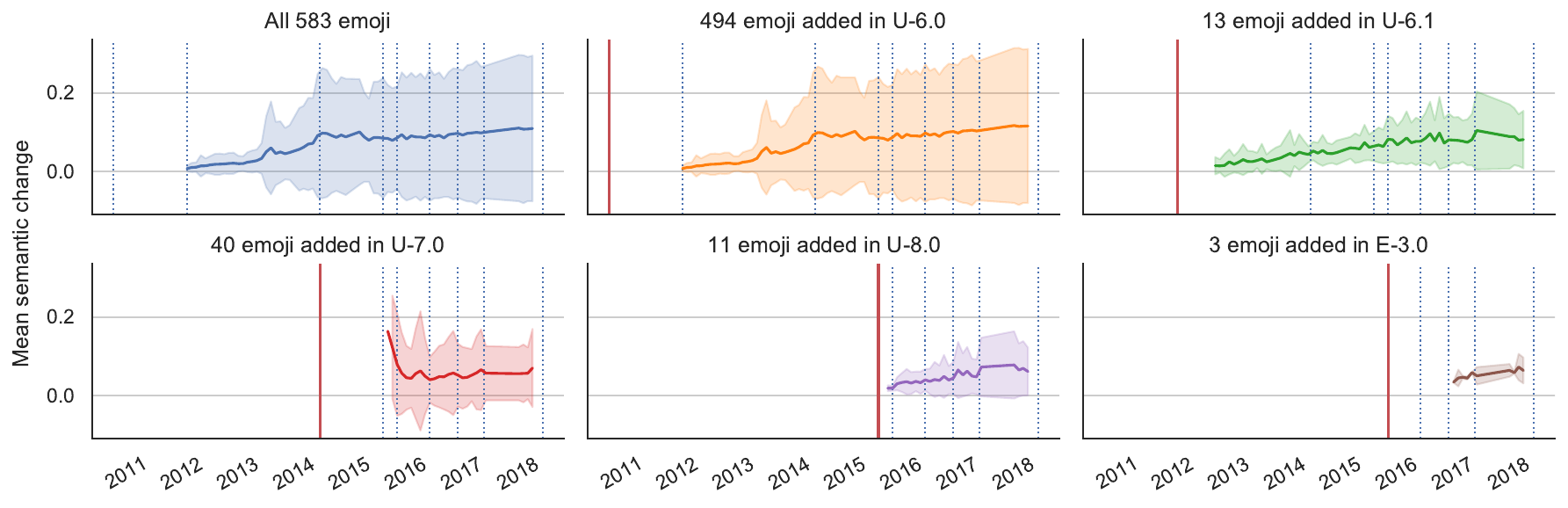}
    \caption{Mean semantic change change score of emoji over time. The standard deviation of the data at each point is shown by the shaded areas. Upper-left plot: all emoji. Others: specific emoji releases. Dashed vertical lines denote addition of new emoji to the Unicode Standard by the Unicode Consortium. The solid vertical line denotes the date of release for that Unicode/Emoji version. First occurrences of emoji lag behind their Unicode addition as they are not immediately implemented by vendors such as Apple and Google.}
    \label{fig:RQ1_overall_change}
\end{figure*}

\subsection{Clustering}\label{methods:clustering}

We can construct a time series using SC at each time point, showing an emoji's change over time. Clustering them, to identify common patterns, proved difficult using standard techniques due to missing data points (some emoji have no embeddings for some months) and time series of differing lengths (emoji can have different anchor points and different end points). Imputation of missing values through linear interpolation can help with the former, but not the latter. One approach would be to focus only on the 125 emoji for which we have complete data, or to separately cluster groups of emoji based on their having the same number of data points. We feel this would limit the scope of our analysis---that one emoji has 65 months of data and another 64 should not warrant separate analyses.

Instead we use a simple technique to generate fixed-length vectors for any emoji. First, we impute missing values using linear interpolation, but only for months between an emoji's first appearance (i.e. the anchor point) and the last month that emoji has data. This prevents us from accidentally ``inventing'' a starting/ending pattern for any emoji.

Next, because the data is highly specific to each emoji, we apply exponential smoothing. This technique slides a window across each time series, fitting that local subset of points to a polynomial by least squares. The new time series has a smoother appearance while retaining the overall characteristics of the original. This makes it more suitable for applying similarity-based algorithms, where we are not necessarily concerned about the finer details of the data. This is a common step in signal processing. We use the Savitzky Golay method \citep{SavitzkyGolay} for smoothing. We use a small sliding window of 5 and a polynomial of degree 3 to avoid over-fitting as this  would exaggerate some of the local characteristics of the time series we wish to filter out: see \citet{HASSANPOUR2012786} for an overview. 

The smoothed data amplifies the general characteristics of the semantic trajectories, but they still have very different scales and start/begin at different points in time. We therefore follow the recommendation of \citet{keogh2003need}---each smoothed time series was individually z-normalised, which scales and shifts a time series such that it has zero mean and zero standard deviation.

Finally, we calculate the similarity of all pairs of emoji timelines using Dynamic Time Warping (DTW) \citep{dtw}. This method allows comparison of time series of different lengths using any distance metric---we use Euclidean distance. For any given emoji, we take the 10 other emoji with the most similar time series, as determined by DTW. After one-hot encoding the names of the most similar emoji to create a feature vector, we use hierarchical clustering to group them. This algorithm is suited to finding unevenly-sized clusters: we do not expect all semantic patterns in emoji to describe uniform numbers of emoji.


This approach is in a similar spirit to \citet{mylonas2004using} who used pairwise distance metrics between individual features of high dimensionality datasets as a form of dimensionality reduction, stating ``the primary aim of clustering algorithms is not to correctly classify data, but rather to identify the patterns that underlie in it and produce clusters of similar data samples. Therefore, `wrong elements' in clusters may be acceptable, as long as the overall cluster correctly describes an existing and meaningful pattern''. This concept is more formally presented as exploiting ultrametric embeddings by \citet{murtagh2008hierarchical} who, again, use the concept of feature-by-feature similarity to find informative subsets of features to perform ``data condensation'' on very high-dimensionality data. The main difference in our method is that we induce a compact form of representation based on similarity of each entire item in our dataset, rather than on the individual features of the items.

\section{Analyses}

We first explore the overall picture of semantic change in emoji, using the techniques described in \Vref{methods:scc}. We focus on 350 emoji with anchor points in 2012, since we have the most data for these. This will allow us to observe the long-term semantic development of a large group of emoji from very close to their release in late 2011 until 2018. We then extend this by looking more closely at individual emoji in this group, to get an understanding of any differences within the group. For these analyses we use SC scores over time, at the aggregate and individual level.

Next, we apply the clustering approach described in \Vref{methods:clustering} to identify more specific patterns of semantic change. For example, do some emoji change very drastically compared to their initial selves as observed in 2012? Perhaps some have essentially fixed semantics. We are interested in quantifying how many patterns there are and how the emoji are distributed across these patterns.

\subsection{An initial inventory of semantic development}

Using the method described in \Vref{methods:scc}, we plot the mean month-to-month SC for all emoji, and then group by the Unicode version in which they first appeared (Figure~\ref{fig:RQ1_overall_change}). 

Considering all emoji together, a general pattern emerges: an initial period where semantics changes very little, followed by a gradual increase in SC scores. Later additions to the standard, though we have less data for them, appear to be following the same general trend as the more established Unicode 6.0 emoji. The standard deviations for SC indicate that this change is not evenly distributed. It is likely to be driven by some subset of emoji. This is especially the case for Unicode 6.0 emoji. Emoji in other releases appear to undergo less semantic change, but again there is a gradual increase in the mean SC scores accompanied by an increase in the standard deviation of those scores.

\begin{figure}
    \centering
    \includegraphics[width=\columnwidth]{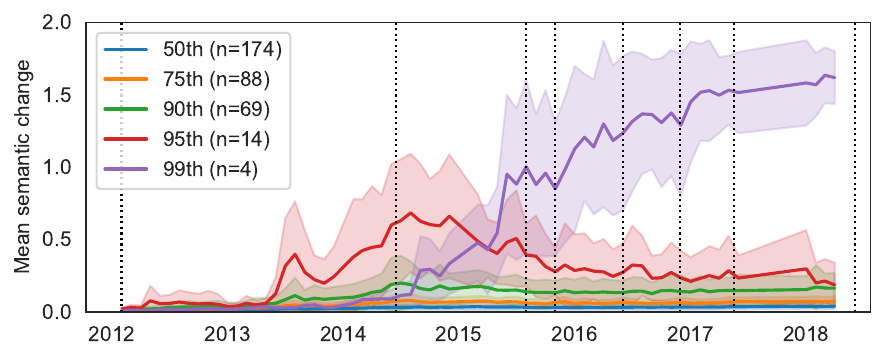}
    \caption{Mean semantic score change over time for Unicode 6.x emoji, grouped by percentile based on standard deviation of their month-to-month semantic change scores. The standard deviation of the data at each point is shown by the shaded areas.}
    \label{fig:RQ2_quantile_change}
\end{figure}

\begin{figure}
    \centering
    \includegraphics[width=\columnwidth]{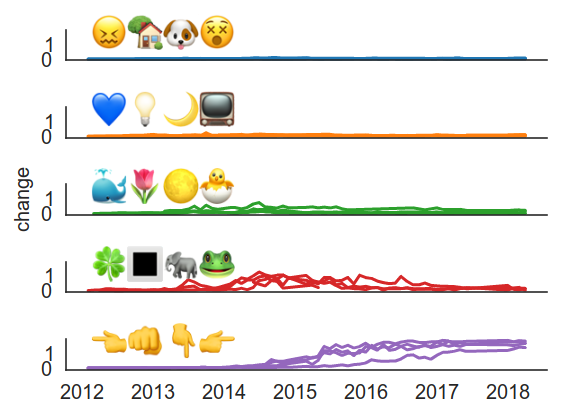}
    \caption{Semantic change over time for 5 random emoji in the 50th (top), 75th, 90th, 95th and 99th (bottom) percentile, based on standard deviation of month-to-month semantic change for Unicode 6.x emoji with 2012 anchor points.}
    \label{fig:RQ2_top5}
\end{figure}

To tease apart the aggregate picture and determine which emoji contribute to the variance in SC, we calculate per-emoji standard deviation of the SC scores and group them into five percentiles. We focus on emoji which have anchor points in 2012---these are all Unicode 6.0 or 6.1 and comprise 348 emoji\footnote{Not 350, as \emoji{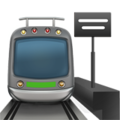} and \emoji{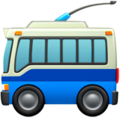} only have one embedding each due to low frequency.} out of a possible total of 507 emoji. \Vref{fig:RQ2_quantile_change} shows the mean SC for these subsets. The variation observed in \Vref{fig:RQ1_overall_change} is due to a very small subset of emoji with a wider range of SC scores over time, rather than all emoji having highly variable SC. This is illustrated very clearly in \Vref{fig:RQ2_top5}, where we show the SC scores for a random sample of each percentile.

The vast majority of emoji, those not in the 95th percentile or above, exhibit very little semantic change over the 6 years, though we do observe some small peaks here and there. The more variable emoji of the 95th percentile appear to exhibit larger, more prolonged, peaks of change. The 99th percentile, just 4 emoji, show significant and sustained change in the second half of their lives.

From these more initial patterns we can begin to build an inventory of emoji semantic development. For most emoji, their semantics is stable from the beginning. Though SC scores may trend upwards over time, this is very slight. For a few emoji, their semantics has highly changed since their first introduction. And in between these two extremes are emoji which, at some point in their life so far, have experienced a rise and fall (to different extents) in their SC scores, suggesting they diverged from their initial meaning but only temporarily.

\subsection{A detailed inventory of semantic development}\label{subsection:detailed}

\begin{figure*}
    \centering
    \includegraphics[width=\linewidth]{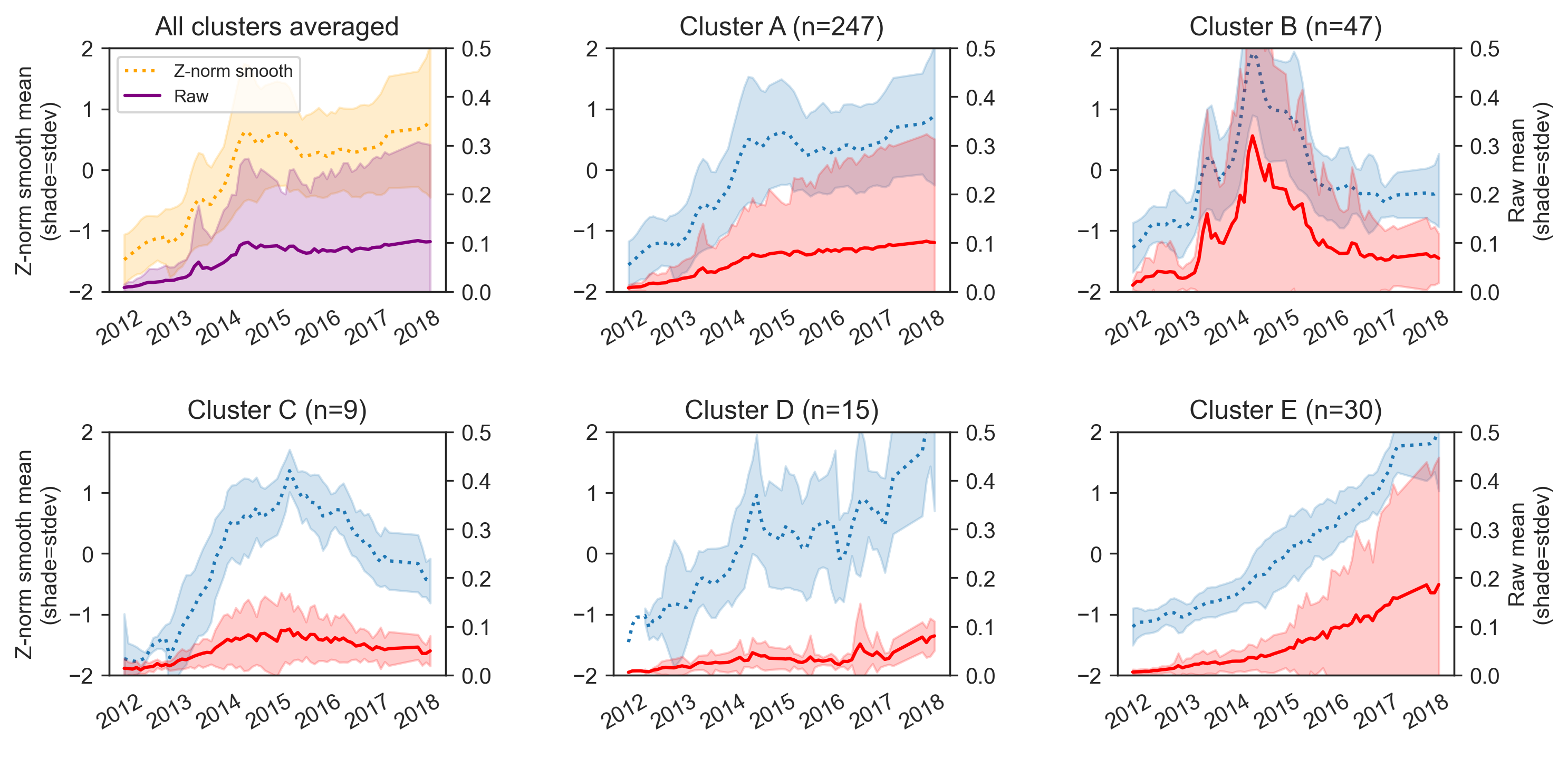}
    \caption{Progression of semantic change in 348 Unicode 6.x emoji, with 2012 anchor points, from 2012 to 2018. Each plot shows a cluster's characteristic shape (dashed lines) and the mean of the actual observed semantic change scores (solid lines). Top left shows the averages over all clusters. The standard deviation of the data at each point is shown by the shaded areas.}
    \label{fig:RQ3_clusters}
\end{figure*}

Our analysis so far has looked at very broad patterns of semantic development. The finding that only a small number of emoji appear to undergo substantive semantic change, and that these changes can happen at different points in time and to different extents (see \Vref{fig:RQ2_top5}), motivates our effort to classify emoji according to their semantic change patterns.

As we are dealing with many emoji, we employ the clustering technique detailed in \Vref{methods:clustering} to automate the process. For each emoji we generate a characteristic shape of the timeline of its monthly SC scores. We compare all pairs of characteristic shapes to find the ten emoji with the most similar timeline shape. These ten emoji are one-hot encoded to create a fixed-length feature vector for use in hierarchical agglomerative clustering. We use the same 348 emoji as in the previous analysis. We identify 5 clusters---this number was chosen manually, such that the mean characteristic shapes of each cluster were visually dissimilar but without there being very few members in each cluster. Results are shown in \Vref{fig:RQ3_clusters}.

The trend for all emoji matches that of \Vref{fig:RQ1_overall_change}---emoji semantics changes gradually before settling down. But again there is a large degree of variation. Within this variation, we see five distinct patterns.

Cluster A (gradually establishing, rank 1/5 with n=247) matches the general trend of the aggregate data---most emoji change relatively slowly, though there is some variation in the extent of this change. Though their semantics has diverged from their initial form to different degrees (as shown by the standard deviation), they are relatively stable.

Cluster B (extreme sudden peak, rank 2/5 with n=47) contains emoji which, overall, remain similar to their initial versions. However, at some point they experiences an extreme but temporary change in their semantics.

Emoji in cluster C (slight gradual peak, rank 5/5 with n=9) also experience a temporary change in semantics but to a far lesser degree than Cluster B emoji. The extent of the changes is also more similar within this cluster, with less variation, as shown by the far lower standard deviation.

Cluster D (static, rank 3/5 with n=15) changed very little for the larger part of their life but may be undergoing change now. These emoji have been otherwise very stable.

Cluster E (not yet established, rank 3/4 with n=30) have been changing from their initial semantics ever since the start. This change appears to be accelerating but to different extents for the emoji in this cluster.

This analysis gives a more nuanced picture of emoji semantic development. Emoji generally have fixed, stable semantics within the time period we observed. The main factor contributing to variation in Cluster A could be due to seasonality---this cluster contains \emoji{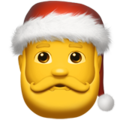} and \emoji{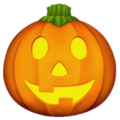} and their SC scores swing dramatically once per year, but on average, are very similar to their 2012 SC scores. See \Vref{fig:frog} for the \emoji{jack-o-lantern}-specific data.


\section{The role of concreteness in emoji semantic change}

Our analyses showed that subsets of the 350 emoji introduced around 2012 underwent different semantic journeys over the next six years. This could be due to the lack of any prescriptive authority (or intent) for emoji, as discussed in \Vref{sec:intro}, but we conjecture that some emoji have more potential for having fixed semantics. 

\begin{figure}
    \centering
    \includegraphics[width=\columnwidth]{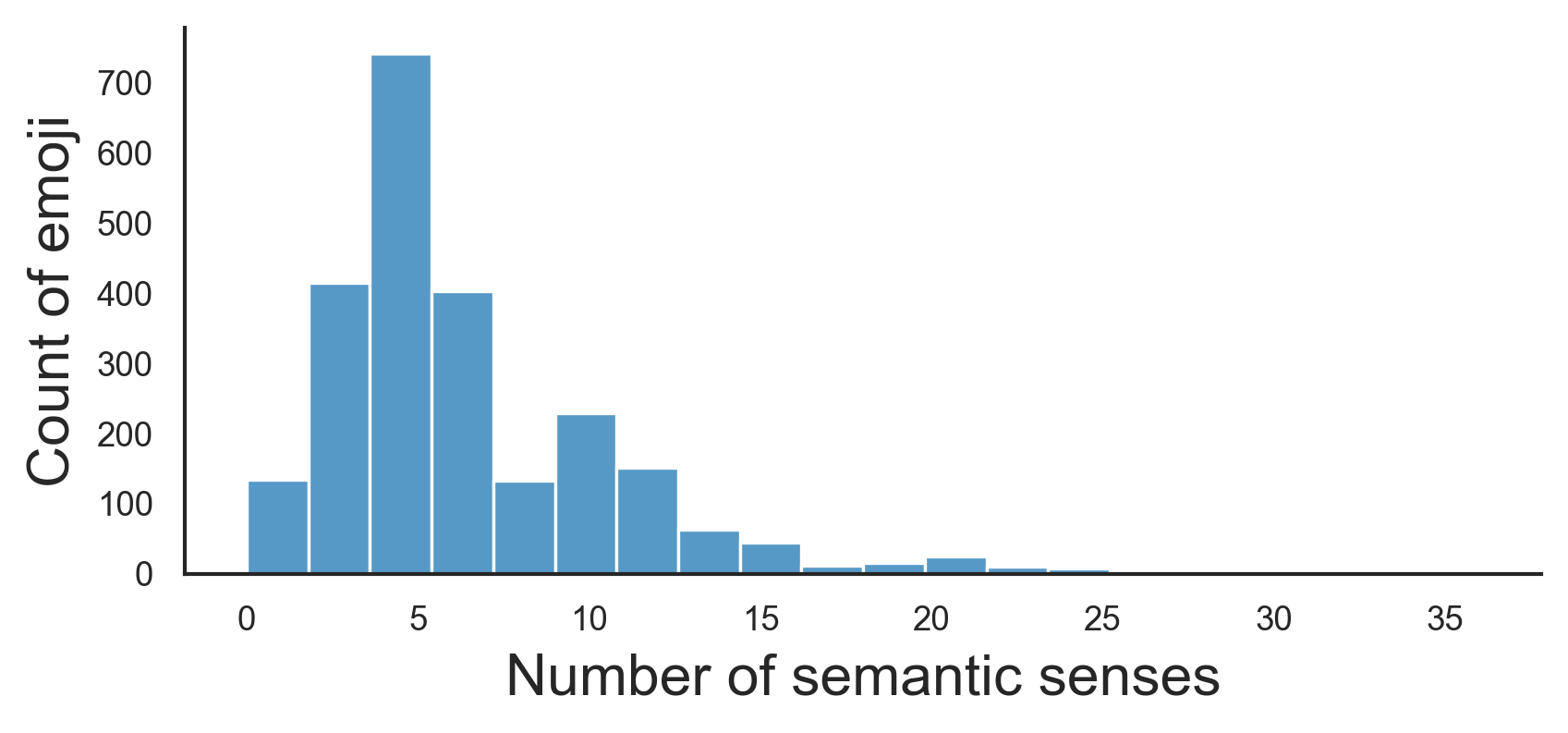}
    \caption{Distribution of the number of senses per emoji, based on EmojiNet data linked to BabelNet senses.}
    \label{fig:senses}
\end{figure}

Previous computational studies on the distributional semantic properties of concrete and abstract words \citep{naumann-etal-2018-quantitative} have found evidence for consistent differences between these two groups. More specifically, \citet{naumann-etal-2018-quantitative} found partial support for the hypothesis that concrete words have significantly less diverse contexts compared to abstract words. Given their specific characteristics, emoji represent an important linguistic category to investigate alongside words with respect to properties along the concrete/abstract scale \cite{wicke}. 
To investigate the relation between concrete versus abstract emoji and their degree of semantic change, we devised a pipeline drawing on EmojiNet \citep{emojineticwsm}, BabelNet \citep{babelnet} and the concreteness dataset by \citet{brysbaert}. As a first step we used EmojiNet to determine how many semantic senses emoji have. The distribution is shown in \Vref{fig:senses}.

We consider an emoji to have undergone a high degree of semantic change if it is in the top 10\% in terms of standard deviation of its month-to-month SC scores. This criterion selects 53 emoji. As EmojiNet provides senses but not lemmas which we can easily look up in the concreteness score dataset, we cross-reference the EmojiNet senses with BabelNet which does provide lemmas. Not all emoji senses appear in the concreteness dataset, however, so we generate scores for 16 emoji in total, with results shown in \Vref{tab:concrete}. The mean concreteness score of these emoji is 4.59 ($\sigma$=0.54), while the full concreteness dataset is 3.04 ($\sigma$=1.04).

We evaluated the null hypothesis that the expected value of the sample of the semantically changed emoji is equal to the mean of the full dataset results, using a two-sided $t$-test. This resulted in a $t$-statistic of 13.52 and a $p$-value $\ll$ 0.01, meaning that that we can reject the null hypothesis. In other words, semantically changed emoji seem to have a statistically significantly different (higher) average concreteness score compared to all emoji. As we discuss in section \ref{sec:discussion}, a comparison between this result and the degree of semantic change of abstract vs. concrete words is an area of further research.

\begin{table}[]
\centering
\resizebox{\columnwidth}{!}{%
\begin{tabular}{@{}cclcclcc@{}}
\cmidrule(r){1-2} \cmidrule(lr){4-5} \cmidrule(l){7-8}
\textbf{Emoji}        & \textbf{Score} &  & \textbf{Emoji}           & \textbf{Score} &  & \textbf{Emoji}        & \textbf{Score} \\ \cmidrule(r){1-2} \cmidrule(lr){4-5} \cmidrule(l){7-8} 
\emoji{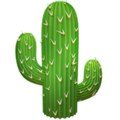}    & 5    &  & \emoji{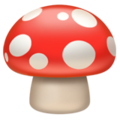} & 4.83 &  & \emoji{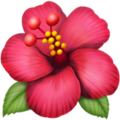} & 4.48                 \\
\emoji{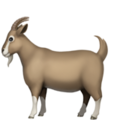}      & 5    &  & \emoji{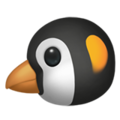}  & 4.8  &  & \emoji{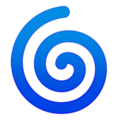}  & 4.48                 \\
\emoji{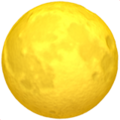} & 5    &  & \emoji{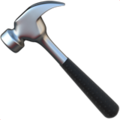}   & 4.77 &  & \emoji{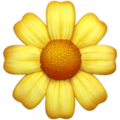}  & 4.26                 \\
\emoji{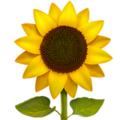} & 5                     &  & \emoji{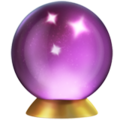} & 4.73                  &  & \emoji{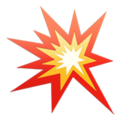} & 3.61                  \\
\emoji{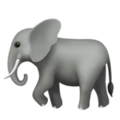}  & 5    &  & \emoji{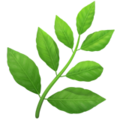}     & 4.57 &  & \emoji{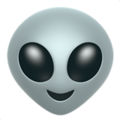}    & 3.52                 \\
\emoji{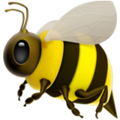}  & 4.86 &  &                      &      &  & \multicolumn{1}{l}{} & \multicolumn{1}{l}{} \\ \cmidrule(r){1-2} \cmidrule(lr){4-5} \cmidrule(l){7-8} 
\end{tabular}%
}
\caption{Concreteness scores of most semantically changed emoji, based on based on the concreteness ratings of \citet{brysbaert}.}
\label{tab:concrete}
\end{table}

\section{Case studies}

We now examine in more detail how specific emoji have changed from 2012 to 2018. We select these to explore different patterns of seasonality (\emoji{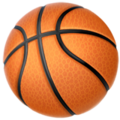} \emoji{jack-o-lantern} \emoji{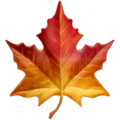}) and fads (\emoji{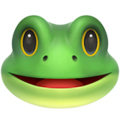}) in emoji semantics, as well as one case (\emoji{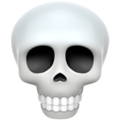}) of a dramatic shift in semantics.

\subsection{\emoji{frog}}

\begin{figure}
    \centering
    \includegraphics[width=\columnwidth]{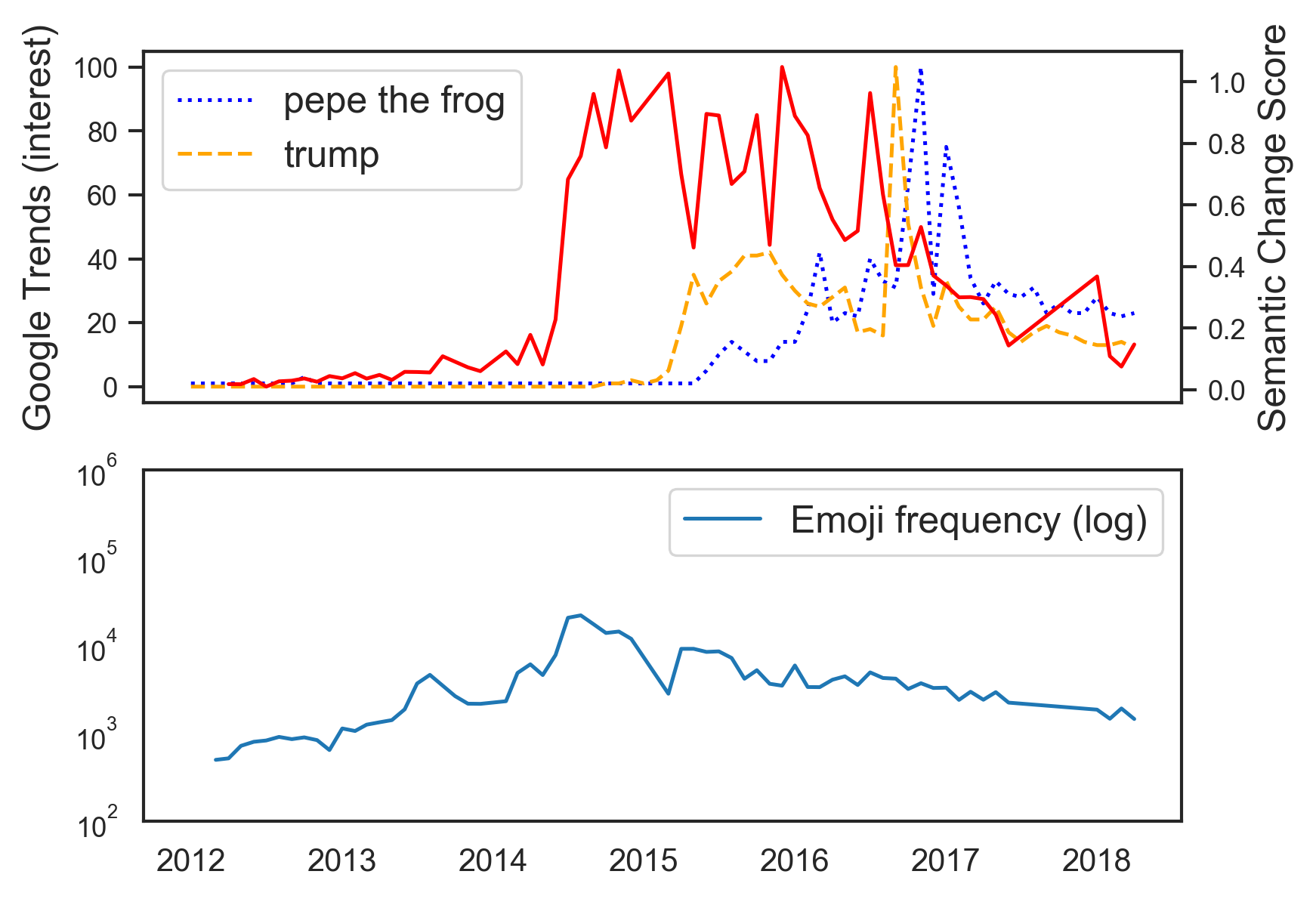}
    \caption{Top: Semantic change (raw, non-interpolated, non-znormed scores) of \emoji{frog} (solid red line). Dashed lines show Google Trends data for ``pepe the frog'' and ``trump''. Bottom: log token frequency per month.}
    \label{fig:frog}
\end{figure}

The frog emoji was assigned to Cluster B in \Vref{subsection:detailed}. These emoji exhibit faddish behaviour, with a sudden intense peak of semantic change. In \Vref{fig:frog} we show the semantic change for \emoji{frog}. According to Emojipedia\footnote{https://emojipedia.org/frog/}, it has been associated with several memes: ``but that's none of my business''\footnote{https://knowyourmeme.com/memes/but-thats-none-of-my-business} and ``Pepe the frog''\footnote{https://knowyourmeme.com/memes/pepe-the-frog}, so we also show Google Trends data on some related terms: ``trump'' and ``pepe the frog''. Both show increased interest around 2016/2017, but it does not appear that this was a direct cause of the overall pattern of change seen in \emoji{frog}---by then it was already very different from its initial version.

The most similar words in 2012 and 2013 were \emph{lizard}, \emph{bunny}, \emph{frog}, but this changed in 2014. Most similar then were \emph{kermit}, \emph{none/nun}, \emph{snitch}, \emph{nvm}, \emph{lowkey}. These are related to the ``but that's none of my business'' meme, but this changed again from 2015 onward when the most similar words become more abusive: the n-word, \emph{bitch}, \emph{hate} all appear along with \emph{pepe}. As of 2017, it was returning not to its 2012 animal-focused meaning, but to the 2014 meme-based meaning as well as taking on a new meaning related to sharing gossip and ``spilling the tea''.







\subsection{\emoji{skull}}

\begin{figure}
    \centering
    \includegraphics[width=\columnwidth]{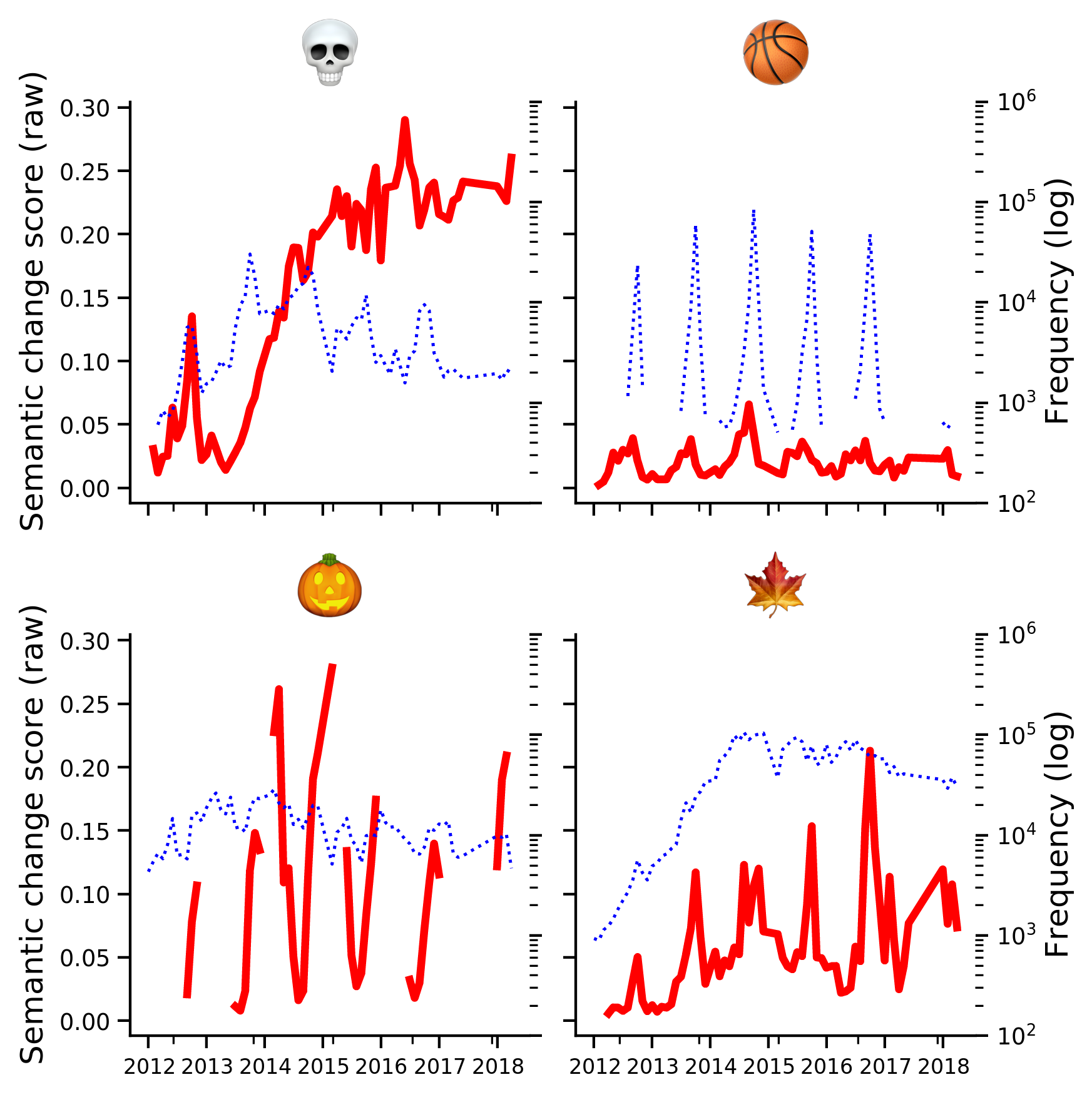}
    \caption{Semantic change patterns (raw, non-interpolated, non-znormed scores - red solid line) and log token frequency (dashed blue line) for four select emoji.}
    \label{fig:four}
\end{figure}

The skull emoji was assigned to Cluster A, the largest group and containing emoji that are generally undergoing gradual change. We selected this one for closer study because it is well known to have a figurative rather than literal meaning and is associated with dying of laughter or embarrassment\footnote{https://emojipedia.org/skull/}. 

This change happened very quickly in 2013, as can be seen in \Vref{fig:four}. The most similar words in 2012 were \emph{zombie}, \emph{corpse}, \emph{bury}, \emph{undead}, \emph{murder}. In 2013, these were joined by versions of \emph{lmao} and entirely replaced by similar terms in 2014.


\subsection{\emoji{basketball}}

Another Cluster A emoji, this has undergone very little change on average but has seen a small temporary change each year so far (\Vref{fig:four}). By examining the monthly most similar words over all years, rather than by the entire year, we observe that the change aligns with the NBA basketball season in the USA, which runs from October to April. Each year from August to October, NBA becomes the most similar term, along with NBA teams and players. At other times of the year, this emoji is used more broadly with terms such as \emph{volleyball}, \emph{soccer} and \emph{softball}.


\subsection{\emoji{jack-o-lantern}}

The jack-o-lantern emoji (\Vref{fig:four}) is also in Cluster A. It is even more seasonal than basketball---so seasonal, in fact, that we have no data for it for half of each year. The nearest neighbours each month are almost all related to Halloween or other holidays, such as Christmas and Thanksgiving. Less similar are more general fantasy and horror terms. It is possible that these are the ``background'' semantics that we would observe if we had set a lower frequency threshold for this emoji.

\subsection{\emoji{maple-leaf}}

The maple leaf emoji (\Vref{fig:four}) is in Cluster D, a small group of emoji which have generally been stable when analysed in terms of their general semantic development. This is true of the maple leaf emoji, on average---it remains similar to its initial version, subject to some seasonal changes.

In this sense the maple leaf emoji is both very similar and very different to the basketball emoji. They both have a seasonal change in meaning but while basketball's two modes are quite similar, the maple leaf's modes are not. From August to November, its most similar words are \emph{autumn}, \emph{winter}, \emph{chilly} and \emph{pumpkin}. From December to July, it is used in connection with cannabis and its most similar terms are \emph{weed}, \emph{smoke}, \emph{blunt} and \emph{stoner}.


\section{Discussion}\label{sec:discussion}

Our analyses establish a general picture of semantic development in emoji, while case studies showed that semantic change can be linked to different kinds of seasonality, or to world events, or a shift from literal to figurative usage. The SC patterns identified here could be further developed into more specific sub-types and give an even more detailed understanding of emoji semantics.

One limitation of this work is that we only considered Unicode 6.x emoji, but it could be extended to others. Another is that our data collection ended in 2018, so we cannot determine the effect of the global pandemic on emoji semantics. For example, the 2020--2021 NBA season was delayed until December 2020---did this have an effect on the seasonality of the basketball emoji?

We expected abstract emoji to undergo more change than concrete emoji, but found the opposite. The visual affordances of emoji may account for this: \emoji{maple-leaf} can denote marijuana because of shared visual similarities, in this case the shape of the leaves. The Lisbon Emoji and Emoticon Database (LEED) \cite{rodrigues2018lisbon} contains subjective norms (including visual complexity and concreteness) for some emoji which could be used to explore the connection between emoji appearance and semantic change. Because our embeddings were trained on English data and LEED norms are based on native Portuguese speakers only, we did not perform this analysis ourselves. However, \citet{rodrigues2018lisbon} make their study materials available through the Open Science Framework, which should make it easy to extend LEED to additional languages.

\alex{As with words, only a minority of emoji exhibit semantic change. Therefore,} a major extension of the work here would be to compare emoji semantic development to that of words. Semantic change in words is often described as an S-shaped curve \cite{feltgen2017frequency}. The semantic development of \emoji{skull} as shown in \Vref{fig:four}, from literal to figurative death, does seem S-shaped. Future work could explore the universality of semantic change. 

Finally, although our analyses cover 6 years of data, it may be too soon to draw any firm conclusions on semantic development in emoji. It may require decades or centuries of additional data to determine the true nature of semantic change in emoji. \alex{However, as stated in the introduction, we are uniquely positioned to study this phenomenon from the very beginning.}

\section{Conclusion}

In this paper, we presented the first longitudinal analysis of emoji semantics. Building on advances in NLP on representation learning and methods for measuring semantic change, we studied how the meaning of individual emoji evolved month by month in six years of Twitter data. Most emoji remained stable, and only a small number of emoji appeared to undergo substantive semantic change in our period of analysis. Our analyses also illustrated variation due to seasonal trends, and how emoji semantics is prone to trends and fads.  We also found that semantically changed emoji seem to have higher average concreteness scores; future work should investigate the reason for this further.

Our study is a first step towards a richer understanding of emoji by taking into account their diachronic development. Our analyses open up many future avenues of exploration that could be further explored using our data, which we make publicly available through an interactive website\footnote{https://semantic-change.emoji-research.com - see Appendix A for more details.} where the raw data can also be downloaded.

\section*{Acknowledgements}

Alexander Robertson was supported by the EPSRC Centre for Doctoral Training in Data Science, funded by the UK Engineering and Physical Sciences Research Council (grant EP/L016427/1) and the University of Edinburgh. Farhana Ferdousi Liza acknowledges the support of the Business and Local Government Data Research Centre (ES/S007156/1) funded by the Economic and Social Research Council (ESRC) for undertaking this work. Scott Hale was supported by the Volkswagen Foundation.

\bibliography{bib}

\section*{Appendix A}

The website https://semantic-change.emoji-research.com/ presents an interactive interface for exploring the data used in this paper. The monthly semantic change scores from 2012 to 2018 are available for all emoji and multiple emoji can be displayed and compared as timelines. In addition, for each emoji the most similar words are shown, aggregated by month or by year, to give a sense of what an emoji's semantics are at different times. This semantic neighbour data can be exported in CSV format for the selected emoji. For users who wish to perform their own analysis, the raw semantic change data and 

\section*{Appendix B}


\begin{figure}[hbt!]
    \centering
    \includegraphics[width=\columnwidth]{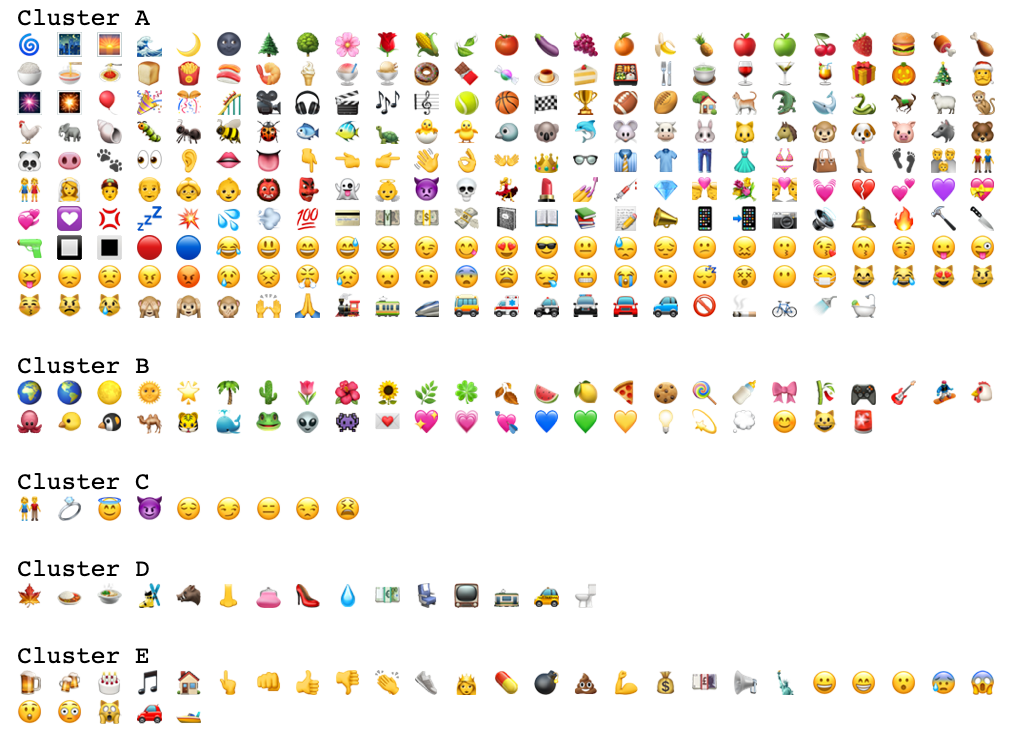}
    \caption{Clustering of 348 Unicode 6.x emoji, based on semantic development from 2012 to 2018.}
    \label{fig:RQ1_emoji}
\end{figure}

\end{document}